\documentclass[runningheads]{llncs}
\usepackage{graphicx}
\usepackage{todonotes}
\usepackage{booktabs}

\usepackage{pgfplots}
\pgfplotsset{width=10cm,compat=1.9}

\usepackage[title]{appendix}

\begin{document}
\title{Classification Betters Regression in Query-based Multi-document Summarisation Techniques for Question Answering}
\subtitle{Macquarie University at BioASQ7b}
%
%
\author{Diego Moll\'a\orcidID{0000-0003-4973-0963} \and
Christopher Jones\orcidID{0000-0002-3491-739X}}
%
%
\institute{Macquarie University, Sydney NSW 2109, Australia\\
\email{Diego.Molla-Aliod@mq.edu.au}\\
\email{Christopher.Jones@students.mq.edu.au}}
\maketitle              
\begin{abstract}
  Task B Phase B of the 2019 BioASQ challenge focuses on biomedical question answering.
  Macquarie University's participation applies query-based multi-document extractive summarisation
  techniques to generate a multi-sentence answer given the question and the set of relevant snippets. In 
  past participation we explored the use of regression approaches using deep
  learning architectures and a simple policy gradient architecture. For the 2019 challenge
  we experiment with the use of classification approaches with and without
  reinforcement learning. In addition, we conduct a correlation analysis between various
  ROUGE metrics and the BioASQ human evaluation scores.

\keywords{Deep learning  \and Reinforcement learning \and Evaluation \and
  Query-based summarisation}
\end{abstract}

\section{Introduction}

The BioASQ Challenge\footnote{\url{http://www.bioasq.org}} includes a question
answering task (Phase B, part B) where the aim is to find the ``ideal answer'' 
--- that is, an answer that would normally be given by a person~\cite{Tsatsaronis:2015}. This is in contrast with most other question answering
challenges where the aim is normally to give an exact answer, usually a
fact-based answer or a list. Given that the answer is based on an input that consists of a
biomedical question and several relevant PubMed
abstracts\footnote{\url{https://www.ncbi.nlm.nih.gov/pubmed/}}, the task can be
seen as an instance of query-based multi-document summarisation.

As in past participation~\cite{Molla:bioasq2017,Molla:bioasq2018}, we wanted to test the use of deep
learning and reinforcement learning approaches for extractive summarisation. In contrast with past years
where the training procedure was based on a regression set up, this year we
experiment with various classification set ups. The main contributions of this
paper are:

\begin{enumerate}
\item We compare classification and regression approaches and show that
  classification produces better results than regression but the quality of the results 
  depends on the approach followed to annotate the data labels. 
\item We conduct correlation analysis between various ROUGE evaluation metrics
  and the human evaluations conducted at BioASQ and show that Precision and F1 correlate better than Recall.
\end{enumerate}

Section~\ref{sec:related} briefly introduces some related work for context.
Section~\ref{sec:clas-reg} describes our classification and regression
experiments. Section~\ref{sec:deep-learning} details our experiments using deep
learning architectures. Section~\ref{sec:rl} explains the reinforcement learning
approaches. Section~\ref{sec:correlation} shows the results of our correlation analysis
between ROUGE scores and human annotations. Section~\ref{sec:runs} lists the specific 
runs submitted at BioASQ~7b. Finally,
Section~\ref{sec:conclusions} concludes the paper.

\section{Related Work}\label{sec:related}

The BioASQ challenge has organised annual challenges on biomedical semantic indexing
and question answering since~2013~\cite{Tsatsaronis:2015}. Every year there has
been a task about semantic indexing (task a) and another about question
answering (task b), and occasionally there have been additional tasks. The tasks defined for 2019 are:

\begin{description}
\item[BioASQ Task 7a:] Large Scale Online Biomedical Semantic Indexing.
\item[BioASQ Task 7b:] Biomedical Semantic QA involving Information Retrieval
  (IR), Question Answering (QA), and Summarisation.
\item[BioASQ MESINESP Task:] Medical Semantic Indexing in Spanish.
\end{description}

BioASQ Task 7b consists of two phases. Phase A provides a biomedical question as an
input, and participants are expected to find relevant concepts from designated
terminologies and ontologies, relevant articles from PubMed, relevant snippets
from the relevant articles, and relevant RDF triples from designated ontologies.
Phase B provides a biomedical question and a list of relevant articles and
snippets, and participant systems are expected to return the exact answers and the ideal
answers. The training data is composed of the test data from all
previous years, and amounts to 2,747 samples. 

There has been considerable research on the use of machine learning approaches
for tasks related to text summarisation, especially on single-document summarisation. Abstractive approaches normally use an encoder-decoder architecture and variants of this architecture incorporate attention~\cite{Rush2015} and
pointer-generator~\cite{See2017}. Recent approaches leveraged the use of pre-trained models~\cite{Hoang2019}. 
Recent extractive approaches to summarisation incorporate recurrent neural networks that model sequences of sentence extractions~\cite{Nallapati2017} and may incorporate an abstractive component and reinforcement learning during the training stage~\cite{Zhang2018}. 
But
relatively few approaches have been proposed for query-based multi-document
summarisation. Table~\ref{tab:techniques} summarises the approaches presented in the proceedings of the 2018 BioASQ challenge.

\begin{table}
  \centering
  
  \caption{Summarisation techniques used in BioASQ 6b for the
    generation of ideal answers. The evaluation result is the human evaluation
    of the best run.}
  \label{tab:techniques}
  \begin{tabular}{lll}
    System & Abstractive Approaches & Extractive Approaches\\
    \midrule
    \cite{Molla:bioasq2018} & (none)  & Regression \& Reinforcement Learning\\
    \cite{Li:bioasq2018}  & Fusion & Maximum Marginal Relevance\\
    \cite{Bhandwaldar:bioasq2018} & (none) & Lexical chains\\
    \cite{Kumar:bioasq2018}     & Fine-tuned Pointer Generator Coverage & Learning to rank\\ 
  \end{tabular}

\end{table}

\section{Classification \emph{vs.} Regression Experiments}
\label{sec:clas-reg}

Our past participation in BioASQ~\cite{Molla:bioasq2017,Molla:bioasq2018} and this paper focus on
extractive approaches to summarisation. Our decision to focus on extractive approaches is based on the observation that a relatively large number of sentences from the input snippets
has very high ROUGE scores, thus suggesting that human annotators had a general
tendency to copy text from the input to generate the target summaries~\cite{Molla:bioasq2017}. Our past participating systems used
regression approaches using the following framework:

\begin{enumerate}
    \item Train the regressor to predict the ROUGE-SU4 F1 score of the input sentence.
    \item Produce a summary by selecting the top $n$ input sentences.
\end{enumerate}

A novelty in the current participation is the introduction of classification approaches using the following framework.

\begin{enumerate}
    \item Train the classifier to predict the target label (``summary'' or ``not summary'') of the input sentence.
    \item Produce a summary by selecting all sentences predicted as ``summary''.
    \item If the total number of sentences selected is less than $n$, select $n$ sentences with higher probability of label ``summary''.
\end{enumerate}

Introducing a classifier makes labelling the training data not trivial, since the target summaries are human-generated and they do not have a perfect mapping to the input sentences. In addition, some samples have multiple reference summaries. \cite{Molla:Louhi2018} showed that different data labelling approaches influence the quality of the final summary, and some labelling approaches may lead to better results than using regression. In this paper we experiment with the following labelling approaches:

\begin{description}
\item[threshold $t$]: Label as ``summary'' all sentences from the input text that have a ROUGE score
  above a threshold $t$.
\item[top $m$]: Label as ``summary'' the $m$ input text sentences with highest ROUGE score.
\end{description}

As in~\cite{Molla:Louhi2018}, The ROUGE score of an input sentence was the ROUGE-SU4 F1 score of the sentence against the set of reference summaries.

We conducted cross-validation experiments using various values of $t$ and $m$. Table~\ref{tab:experiments} shows the results for the best
values of $t$ and $m$ obtained. The regressor and
classifier used Support Vector Regression (SVR) and Support Vector Classification
(SVC) respectively. To enable a fair comparison we used the same input features in all systems. These input features combine information from the question and the input sentence and are shown in Fig.~\ref{fig:SVCfeatures}. The features are based on~\cite{Malakasiotis2015}, and are the same as in~\cite{Molla:bioasq2017}, plus the addition of the position of the input snippet. The best SVC and SVR parameters were determined by grid search. 

\begin{figure}
    \centering
\begin{minipage}{10cm}
    \begin{itemize}
\item $tf.idf$ vector of the candidate sentence.
\item Cosine similarity between the $tf.idf$ vector of the
  question and the $tf.idf$ vector of the candidate sentence.
\item The largest cosine similarity between the $tf.idf$ vector of
  candidate sentence and the $tf.idf$ vector of each of the snippets
  related to the question. 
\item Cosine similarity between the sum of word2vec embeddings of the
  words in the question and the word2vec embeddings of the words in
  the candidate sentence. We used
  vectors of dimension 200 pre-trained using PubMed documents provided by the organisers of BioASQ.
\item Pairwise cosine similarities between the words of the question
  and the words of the candidate sentence. We used word2vec to compute the word
  vectors. We then computed the pairwise cosine
  similarities and selected the following features:
  \begin{itemize}
  \item The mean, median, maximum, and minimum of all pairwise cosine similarities.
  \item The mean of the 2 highest, mean of the 3 highest, mean of the
    2 lowest, and mean of the 3 lowest.
  \end{itemize}
\item Weighted pairwise cosine similarities where the weight was the $tf.idf$ of the word.
\end{itemize}
\end{minipage}
    \caption{Features used in the SVC and SVR experiments.}
    \label{fig:SVCfeatures}
\end{figure}

Preliminary experiments showed a relatively high number of cases where the classifier did not classify any of the input sentences as ``summary''. To solve this problem, and as mentioned above, the summariser used in Table~\ref{tab:experiments} introduces a backoff step that extracts the $n$ sentences with highest predicted values when the summary has less than $n$ sentences. The value of $n$ is as reported in our prior work and shown in Table~\ref{tab:n}.

\begin{table}[]
    \centering
    \caption{Number of sentences returned by the regression-based summarisers and the backoff step of the classification-based summarisers, for each question type}
    \label{tab:n}
    \begin{tabular}{ccccc}
    & \textbf{Summary} & \textbf{Factoid} & \textbf{Yesno} & \textbf{List} \\
    \midrule
    \textbf{n}     &  6 & 2 & 2 & 3\\
    \end{tabular}
\end{table}

The results confirm \cite{Molla:Louhi2018}'s finding that classification outperforms
regression. However, the actual choice of optimal labelling scheme was different: whereas in~\cite{Molla:Louhi2018} the optimal labelling was based on a labelling
threshold of 0.1, our experiments show a better result when using the top 5 sentences as the target summary. The reason for this difference might be the fact that~\cite{Molla:Louhi2018} used all sentences from the abstracts of the relevant PubMed
articles, whereas we use only the snippets as the input to our summariser. Consequently, the number of input
sentences is now much smaller. We therefore report the results of using the labelling schema of top~5
snippets in all subsequent classifier-based experiments of this paper.


  \tikzstyle barchart=[fill=black!20,draw=black]
  \tikzstyle errorbar=[very thin,draw=black!75]
  \tikzstyle sscale=[very thin,draw=black!75]

\newcommand{\mybar}[2]{
  \begin{minipage}[c]{5.5cm}
   \begin{tikzpicture}
    \draw (0cm,0cm) (5.5,0.5);
  \draw[barchart] (0,0.152) rectangle (#1,0.438);
  \draw[errorbar] (#1-#2,0.295) -- (#1+#2,0.295);
  \draw[errorbar] (#1-#2,0.333) -- (#1-#2,0.258);
  \draw[errorbar] (#1+#2,0.333) -- (#1+#2,0.258);
   \end{tikzpicture}
  \end{minipage}
}

\newcommand{\myscale}{
\hspace{-3pt}%
 \begin{minipage}[c]{5.5cm}
   \begin{tikzpicture}
    \draw (0cm,0cm) (5.5,0.3);
    \draw[sscale] (0,0.3) -- (5,0.3);
    \draw[sscale] (0,0.3) -- (0,0.4);
    \draw[sscale] (0,0) node[text width=0pt, text height=0pt, font=\footnotesize] {0.22};
    \draw[sscale] (1,0.3) -- (1,0.4);
    \draw[sscale] (1,0) node[text width=0pt, text height=0pt, font=\footnotesize] {0.23};
    \draw[sscale] (2,0.3) -- (2,0.4);
    \draw[sscale] (2,0) node[text width=0pt, text height=0pt, font=\footnotesize] {0.24};
    \draw[sscale] (3,0.3) -- (3,0.4);
    \draw[sscale] (3,0) node[text width=0pt, text height=0pt, font=\footnotesize] {0.25};
    \draw[sscale] (4,0.3) -- (4,0.4);
    \draw[sscale] (4,0) node[text width=0pt, text height=0pt, font=\footnotesize] {0.26};
    \draw[sscale] (5,0.3) -- (5,0.4);
    \draw[sscale] (5,0) node[text width=0pt, text height=0pt, font=\footnotesize] {0.27};
   \end{tikzpicture}
  \end{minipage} 
  }

\begin{table}[]
    \centering
    \caption{Regression vs. classification approaches measured using
      ROUGE SU4 F-score under 10-fold cross-validation. The table shows the mean
      and standard deviation across the folds. ``firstn'' is a baseline that selects the first n sentences. SVR and SVC are described in Section~\ref{sec:clas-reg}. NNR and NNC are described in Section~\ref{sec:deep-learning}.}
    \label{tab:experiments}
    \begin{tabular}{llr@{ $\pm$ }ll}
         Method & Labelling & \multicolumn{2}{c}{ROUGE-SU4 F1} \\
         & &Mean&1 stdev\\
         \cmidrule{1-4}
      firstn & & 0.252 &  0.015 & \mybar{3.2}{1.5}\\
         \cmidrule{1-4}
         SVR & SU4 F1 & 0.239 &  0.009 & \mybar{1.9}{0.9}\\
         SVC & threshold 0.2 & 0.240 & 0.012 & \mybar{2}{1.2}\\
      SVC & top 5 & 0.253 & 0.013 & \mybar{3.3}{1.3}\\
      \cmidrule{1-4}
      NNR & SU4 F1 & 0.254 & 0.013 & \mybar{3.4}{1.3}\\
      NNC & SU4 F1 & 0.257 & 0.012 & \mybar{3.7}{1.2}\\
      NNC & top 5 & 0.262 & 0.012 & \mybar{4.2}{1.2}\\
      \cmidrule{1-4}
 \multicolumn{4}{l}{} & \myscale\\
    \end{tabular}
\end{table}

\section{Deep Learning Models}\label{sec:deep-learning}

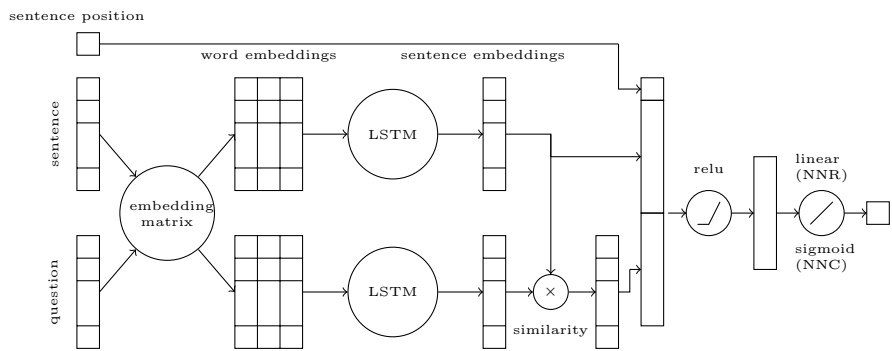
\begin{figure}
  \centering
    \begin{tikzpicture}[scale=0.3]
    \tiny
    \draw (0,0) rectangle (1,5) (0,1) -- (1,1) (0,3) -- (1,3) (0,4) -- (1,4);
    \draw (0,-7) rectangle (1,-2) (0,-6) -- (1,-6) (0,-4) -- (1,-4) (0,-3) -- (1,-3);
    \draw (-1,2.5) node[rotate=90] {sentence};
    \draw (-1,-4.5) node[rotate=90] {question};

    \draw (4,-1) node [circle,draw,align=center,text width=1cm] (em) {embedding matrix};
    \draw (8.5,6) node {word embeddings};
    \draw (7,0) rectangle (10,5) (7,1) -- (10,1) (7,3) -- (10,3) (7,4) -- (10,4) (8,0) -- (8,5) (9,0) -- (9,5);
    \draw (7,-7) rectangle (10,-2) (7,-6) -- (10,-6) (7,-4) -- (10,-4) (7,-3) -- (10,-3) (8,-7) -- (8,-2) (9,-7) -- (9,-2);

    \draw[->] (1,2.5) -- (em);
    \draw[->] (1,-4.5) -- (em);

    \draw[->] (em) -- (7,2.5);
    \draw[->] (em) -- (7,-4.5);
    \draw (14,2.5) node [circle,draw,align=center,text width=1cm] (sr) {LSTM};
    \draw (14,-4.5) node [circle,draw,align=center,text width=1cm] (qr) {LSTM};
    \draw (18,6) node {sentence embeddings};
    \draw (18,0) rectangle (19,5) (18,1) -- (19,1) (18,3) -- (19,3) (18,4) -- (19,4);
    \draw (18,-7) rectangle (19,-2) (18,-6) -- (19,-6) (18,-4) -- (19,-4) (18,-3) -- (19,-3);

    \draw[->] (10,2.5) -- (sr);
    \draw[->] (sr) -- (18,2.5);
    \draw[->] (10,-4.5) -- (qr);
    \draw[->] (qr) -- (18,-4.5);

    \draw (21,-4.5) node [circle,draw] (t) {$\times$};
    \draw (23,-7) rectangle (24,-2) (23,-6) -- (24,-6) (23,-4) -- (24,-4) (23,-3) -- (24,-3);

    \draw[->] (19,2.5) -| (t);
    \draw[->] (19,-4.5) -- (t);

    \draw[->] (21,2.5) |- (25,1.5);
    \draw[->] (t) -- (23,-4.5);
    \draw (24,-4.5) -| (24.5,-3.5);
    \draw[->] (24.5,-3.5) -- (25,-3.5);

    \draw (21,-6.2) node {similarity};

    \draw (25,-1) rectangle (26,4);
    \draw (25,-1) rectangle (26,-6);
    \draw (28,-1) circle[radius=1] (27.5,-1.5) -- (28,-1.5) -- (28.5,-0.5);
    \draw (28,1) node {relu};

    \draw (30,-3.5) rectangle (31,1.5);

    \draw[->] (26.2,-1) -- (27,-1);
    \draw[->] (29,-1) -- (30,-1);

    \draw (33,-1) circle[radius=1] (32.5,-1.5) -- (33.5,-0.5);
    \draw (35,-1.5) rectangle (36,-0.5);
    \draw (33.5,1) node[text width=1cm] {linear (NNR)};
    \draw (33.5,-3) node[text width=1cm] {sigmoid (NNC)};

    \draw[->] (31,-1) -- (32,-1);
    \draw[->] (34,-1) -- (35,-1);
    
    \draw (0,6) rectangle (1,7);
    \draw (1,6.5) -- (24,6.5);
    \draw[->] (24,6.5) |- (25,4.5);
    \draw (25,5) rectangle (26,4);
    \draw (0,7.7) node {sentence position};
  \end{tikzpicture}
  \caption{Architecture of the neural classification and regression systems. A matrix of pre-trained word embeddings (same pre-trained vectors as in Fig.~\ref{fig:SVCfeatures}) is used to find the embeddings of the words of the input sentence and the question. Then, LSTM chains are used to generate sentence embeddings --- the weights of the LSTM chains of input sentence and question are not shared. Then, the sentence position is concatenated to the sentence embedding and the similarity of sentence and question embeddings, implemented as a product. A final layer predicts the label of the sentence.}
  \label{fig:architecture}
\end{figure}

Based on the findings of Section~\ref{sec:clas-reg}, we apply minimal changes
to the deep learning regression models of~\cite{Molla:bioasq2018} to convert them to classification
models. In particular, we add a sigmoid activation to the final layer, and
use cross-entropy as the loss function.\footnote{We also changed the platform from
  TensorFlow to the Keras API provided by TensorFlow.} The complete architecture is shown in
Fig.~\ref{fig:architecture}.

The bottom section of Table~\ref{tab:experiments} shows the results of several
variants of the neural architecture. The table includes a neural regressor (NNR)
and a neural classifier (NNC). The neural classifier is trained in two set ups:
``NNC top 5'' uses classification labels as described in Section~\ref{sec:clas-reg}, and ``NNC SU4 F1'' uses
the regression labels, that is, the ROUGE-SU4 F1 scores of each sentence. Of
interest is the fact that ``NNC SU4 F1'' outperforms the neural regressor. We have not explored this
further and we presume that the relatively good results are due to the fact that
ROUGE values range between 0 and 1, which matches the full range of probability
values that can be returned by the sigmoid activation of the classifier final layer. 

Table~\ref{tab:experiments} also shows the
standard deviation across the cross-validation folds. Whereas this standard
deviation is fairly large compared with the differences in results, in general
the results are compatible with the top part of the table and prior work
suggesting that classification-based approaches improve over regression-based
approaches. 

\section{Reinforcement Learning}\label{sec:rl}

We also experiment with the use of reinforcement learning
techniques. Again these experiments are based on~\cite{Molla:bioasq2018}, who uses 
REINFORCE to train a global policy. The policy predictor uses a
simple feedforward network with a hidden layer. 

The results reported by~\cite{Molla:bioasq2018} used ROUGE Recall and
indicated no improvement with respect to deep learning architectures. Human evaluation results are preferable over ROUGE but these
were made available after the publication of the paper. When comparing the ROUGE and human evaluation results (Table~\ref{tab:Bioasq6b}), we observe an inversion of the results. In particular, the reinforcement learning approaches (RL) of~\cite{Molla:bioasq2018} receive 
good human evaluation results, and as a matter of fact they are the best of our runs in two of the batches. In contrast, the regression systems (NNR) fare relatively poorly.
Section~\ref{sec:correlation} expands on the comparison between the ROUGE and
human evaluation scores.

\begin{table}
  \centering
  \caption{Results of ROUGE-SU4 Recall (R) and human (H) evaluations on BioASQ
    6b runs, batch 5. The human evaluation shows the average of all human
    evaluation metrics.}
  \label{tab:Bioasq6b}
  \begin{tabular}{llrrrrrrrrrr}
    Run & System & \multicolumn{2}{c}{Batch 1} & \multicolumn{2}{c}{Batch 2} & \multicolumn{2}{c}{Batch 3} & \multicolumn{2}{c}{Batch 4} & \multicolumn{2}{c}{Batch 5}\\
    & & R & H & R & H & R & H & R & H & R & H\\ 
    \midrule
    MQ-1 & First $n$ & 0.46 & 3.91 & 0.50 & \textbf{4.01} & 0.45 & \textbf{4.06} & 0.51 & 4.16 & 0.59 & 4.05\\ 
    MQ-2 & Cosine & 0.52 & \textbf{3.96} & 0.50 & 3.97 & 0.45 & 3.97 & 0.53 & 4.15 & 0.59 & 4.06\\
    MQ-3 & SVR & 0.49 & 3.87 & 0.51 & 3.96 & 0.49 & \textbf{4.06} & 0.52 & 4.17 & 0.62 & 3.98\\
    MQ-4 & NNR & \textbf{0.55} & 3.85 & \textbf{0.54} & 3.93 & \textbf{0.51} & 4.05 & \textbf{0.56} & \textbf{4.19} & \textbf{0.64} & 4.02\\
    MQ-5 & RL & 0.38 & 3.92 & 0.43 & \textbf{4.01} & 0.38 & 4.04 & 0.46 & 4.18 & 0.52 & \textbf{4.14}\\
  \end{tabular}
\end{table}

Encouraged by the results of Table~\ref{tab:Bioasq6b}, we decided to continue with our experiments with
reinforcement learning. We use the same features as in~\cite{Molla:bioasq2018}, namely
the length (in number of sentences) of the summary generated so far, plus the
$tf.idf$ vectors of the following:

\begin{enumerate}
\item Candidate sentence;
\item Entire input to summarise;
\item Summary generated so far;
\item Candidate sentences that are yet to be
  processed; and
\item Question.
\end{enumerate}

The reward used by REINFORCE is the ROUGE value of the summary generated by the
system. Since~\cite{Molla:bioasq2018} observed a difference between the ROUGE values of
the Python implementation of ROUGE and the original Perl version (partly because
the Python implementation does not include ROUGE-SU4), we compare the performance
of our system when trained with each of them. 
Table~\ref{tab:rl-experiments} summarises some of our experiments. We ran the
version trained on Python ROUGE once, and the version trained on Perl twice. The
two Perl runs have different results, and one of them clearly outperforms the
Python run. However, given the differences of results between the two Perl runs we
advice to re-run the experiments multiple times and obtain the mean and
standard deviation of the runs before concluding whether there is any
statistical difference between the results. But it seems that there may be an improvement of the
final evaluation results when training on the Perl ROUGE values, presumably
because the final evaluation results are measured using the Perl implementation of ROUGE.

\begin{table}
  \centering
  \caption{Experiments using Perl and Python versions of ROUGE. The Python
    version used the average of ROUGE-2 and ROUGE-L, whereas the Perl version
    used ROUGE-SU4.}
  \label{tab:rl-experiments}
  \begin{tabular}{lrr}
    Training on & Python ROUGE & Perl ROUGE\\
    \midrule
    Python implementation & 0.316 & 0.259\\
    Perl implementation 1 & 0.287 & 0.238\\
    Perl implementation 2 & 0.321 & 0.274\\
  \end{tabular}
\end{table}

We have also tested the use of word embeddings instead of $tf.idf$ as input features to the policy model, while
keeping the same neural architecture for the policy (one hidden layer using the same
number of hidden nodes). In particular, we use the mean of word embeddings using 100 and
200 dimensions. These word embeddings were pre-trained using word2vec on PubMed documents provided by the organisers of BioASQ, as we did for the architectures described in previous sections. The results, not shown in the paper, indicated no major improvement, and re-runs of the experiments showed different results on different runs. Consequently, our submission to BioASQ included the original system using $tf.idf$ as input features in all batches but batch~2, as described in Section~\ref{sec:runs}.

\section{Evaluation Correlation Analysis}\label{sec:correlation}

As mentioned in Section~\ref{sec:rl}, there appears to be a large discrepancy between ROUGE Recall and the human evaluations.
This section describes a correlation analysis between human and ROUGE evaluations
using the runs of all participants to all previous BioASQ challenges that included human evaluations (Phase B, ideal
answers). The human evaluation results were scraped from the BioASQ Results
page, and the ROUGE results were kindly provided by the organisers. We compute the 
correlation of each of the ROUGE metrics (recall,
precision, F1 for ROUGE-2 and ROUGE-SU4) against the average of the human
scores. The correlation metrics are Pearson, Kendall, and a revised Kendall
correlation explained below.

The Pearson correlation between two variables is computed as the covariance of the 
two variables divided by the
product of their standard deviations. This correlation is a good indication of a
linear relation between the two variables, but may not be very effective when
there is non-linear correlation.

The Spearman rank correlation and the Kendall rank correlation are two of the most popular among metrics that aim to detect non-linear correlations. The Spearman rank correlation between two variables can be computed as the Pearson correlation between the rank
values of the two variables, whereas the Kendall rank correlation measures the
ordinal association between the two variables using Equation~\ref{eq:kendall}.

\begin{equation}
  \label{eq:kendall}
  \tau=\frac{(\hbox{number of concordant pairs})-(\hbox{number of discordant pairs})}{n(n-1)/2}
\end{equation}

It is useful to account for the fact that the results are
from 28 independent sets (3 batches in BioASQ 1 and 5 batches each year between BioASQ 2 and BioASQ 6). We therefore also compute a revised Kendall rank correlation measure that only
considers pairs of variable values within the same set. The revised
metric is computed using Equation~\ref{eq:kendall2}, where $S$ is the list of different sets.

\begin{equation}
  \label{eq:kendall2}
  \tau'=\frac{\sum_{i\in S}\left[(\hbox{number of concordant pairs})_i-(\hbox{number of discordant pairs})_i\right]}{\sum_{i\in S}\left[n_i(n_i-1)/2\right]}
\end{equation}

Table~\ref{tab:correlation} shows the results of all correlation metrics. Overall, ROUGE-2 and ROUGE-SU4
give similar correlation values but ROUGE-SU4 is marginally better. Among
precision, recall and F1, both precision and F1 are similar, but precision gives a
better correlation. Recall shows poor correlation, and virtually no correlation when using the revised
Kendall measure. For reporting the evaluation of results, it will be therefore
more useful to use precision or F1. However, given the small difference between
precision and F1, and given that precision may favour short summaries when used
as a function to optimise in a machine learning setting (e.g. using
reinforcement learning), it may be best to use F1 as the metric to optimise.

\begin{table}
  \centering
  \caption{Correlation analysis of evaluation results}
  \label{tab:correlation}
  \begin{tabular}{l@{~~~}r@{~~~~~}r@{~~~~~}r@{~~~~~}r}
    Metric & Pearson & Spearman & Kendall & Revised Kendall\\
    \midrule
    ROUGE-2 precision &  0.61 & 0.78 & 0.58 &  0.73\\
    ROUGE-2 recall    &  0.41 & 0.24 & 0.16 & -0.01\\
    ROUGE-2 F1        &  0.62 & 0.68 & 0.49 &  0.42\\
    ROUGE-SU4 precision & 0.61 & 0.79 & 0.59 &  0.74\\
    ROUGE-SU4 recall    & 0.40 & 0.20 & 0.13 & -0.02\\
    ROUGE-SU4 F1        & 0.63 & 0.69 & 0.50 &  0.43\\
  \end{tabular}
\end{table}

\begin{figure}
    \centering
\includegraphics[width=6cm]{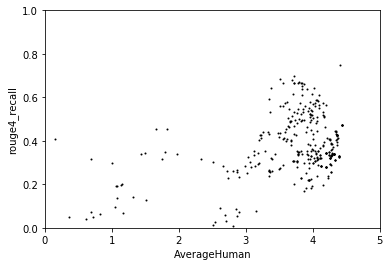}    
\includegraphics[width=6cm]{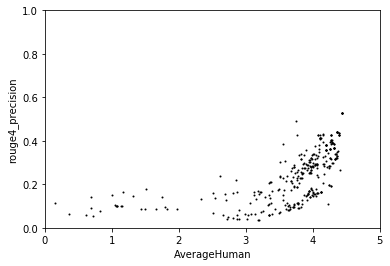}    
\includegraphics[width=6cm]{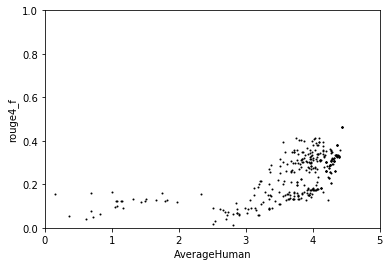}    
\caption{Scatterplots of ROUGE SU4 evaluation metrics against the average human evaluations.}
    \label{fig:scatterplots}
\end{figure}

Fig.~\ref{fig:scatterplots} shows the scatterplots of ROUGE-SU4 recall, precision and F1 with respect to the average human evaluation\footnote{The scatterplots of ROUGE-2 are very similar to those of ROUGE-SU4}. We observe that the relation between ROUGE and the human evaluations is not linear, and that Precision and F1 have a clear correlation.

\section{Submitted Runs}\label{sec:runs}

Table~\ref{tab:bioasq7b} shows the results and details of the runs submitted to BioASQ. The
table uses ROUGE-SU4 Recall since this is the metric available at the time of writing this paper. 
However, note that, as explained in Section~\ref{sec:correlation}, these results
might differ from the final human evaluation results. Therefore we do not
comment on the results, other than observing that the ``first $n$''
baseline produces the same results as the neural regressor. As mentioned in
Section~\ref{sec:clas-reg}, the labels used for the classification experiments
are the 5 sentences with highest ROUGE-SU4 F1 score.

\begin{table}
  \centering
  \caption{Runs submitted to BioASQ 7b}
  \label{tab:bioasq7b}
  \begin{tabular}{lllr}
    Batch & Run & Description & ROUGE-SU4 R\\
    \midrule
    1 & MQ1 & First $n$ & 0.4741\\
          & MQ2 & SVC & 0.5156\\
          & MQ3 & NNR batchsize=4096 & 0.4741\\
          & MQ4 & NNC batchsize=4096 & 0.5214\\
          & MQ5 & RL tf.idf \& Python ROUGE & 0.4616\\
    \midrule
    2 & MQ1 & First $n$ & 0.5113\\
          & MQ2 & SVC & 0.5206\\
          & MQ3 & NNR batchsize=4096 & 0.5113\\
          & MQ4 & NNC batchsize=4096 & 0.5337\\
          & MQ5 & RL embeddings 200 \& Python ROUGE & 0.4787\\
    \midrule
    3 & MQ1 & First $n$ & 0.4263\\
          & MQ2 & SVC & 0.4512\\
          & MQ3 & NNR batchsize=4096 & 0.4263\\
          & MQ4 & NNC batchsize=4096 & 0.4782\\
          & MQ5 & RL tf.idf \& Python ROUGE & 0.4189\\
    \midrule
    4 & MQ1 & First $n$ & 0.4617\\
          & MQ2 & SVC & 0.4812\\
          & MQ3 & NNR batchsize=1024 & 0.4617\\
          & MQ4 & NNC batchsize=1024 & 0.5246\\
          & MQ5 & RL tf.idf \& Python ROUGE & 0.3940\\
    \midrule
    5 & MQ1 & First $n$ & 0.4952\\
          & MQ2 & SVC & 0.5024\\
          & MQ3 & NNR batchsize=1024 & 0.4952\\
          & MQ4 & NNC batchsize=1024 & 0.5070\\
          & MQ5 & RL tf.idf \& Perl ROUGE & 0.4520\\
    \midrule
    
  \end{tabular}
\end{table}

\section{Conclusions}\label{sec:conclusions}

Macquarie University's participation in BioASQ 7 focused on the task of
generating the ideal answers. The runs use
query-based extractive techniques and we experiment with classification,
regression, and reinforcement learning approaches. At the time of writing there
were no human evaluation results, and based on ROUGE-F1 scores under cross-validation on the training data
we observed that
classification approaches outperform regression approaches. We experimented
with several approaches to label the individual sentences for the classifier and
observed that the optimal labelling policy for this task differed from prior
work.

We also observed poor correlation between ROUGE-Recall and human evaluation
metrics and suggest to use alternative automatic evaluation metrics with better
correlation, such as ROUGE-Precision or ROUGE-F1. Given the nature of
precision-based metrics which could bias the system towards returning short
summaries, ROUGE-F1 is probably more appropriate when using at development time,
for example for the reward function used by a reinforcement learning system.

Reinforcement learning gives promising results, especially in human evaluations
made on the runs submitted to BioASQ 6b. This year we introduced very small changes to the runs 
using reinforcement learning,
and will aim to explore more complex reinforcement learning strategies and more complex neural models
in the policy and value estimators.

%
%
%
\bibliographystyle{splncs04}
\bibliography{mybibliography}

\begin{thebibliography}{10}
\providecommand{\url}[1]{\texttt{#1}}
\providecommand{\urlprefix}{URL }
\providecommand{\doi}[1]{https://doi.org/#1}

\bibitem{Bhandwaldar:bioasq2018}
Bhandwaldar, A., Charlotte, U.N.C., Charlotte, U.N.C.: {UNCC QA : A Biomedical
  Question Answering System}. In: Proceedings BioASQ Workshop at EMNLP 2018.
  pp. 66--71 (2018)

\bibitem{Hoang2019}
Hoang, A., Bosselut, A., Celikyilmaz, A., Choi, Y.: {Efficient Adaptation of
  Pretrained Transformers for Abstractive Summarization}. Arxiv pre-print
  1906.00138  (may 2019)

\bibitem{Molla:Louhi2018}
Kaur, M., Moll{\'{a}}, D.: {Supervised Machine Learning for Extractive Query
  Based Summarisation of Biomedical Data}. In: Proc. Louhi 2018 (2018)

\bibitem{Li:bioasq2018}
Li, Y., Gekakis, N., Chandu, K.R., Nyberg, E.: {Extraction Meets Abstraction :
  Ideal Answer Generation for Biomedical Questions}. In: Proceedings BioASQ
  Workshop at EMNLP 2018. pp. 57--65 (2018)

\bibitem{Malakasiotis2015}
Malakasiotis, P., Archontakis, E., Androutsopoulos, I.: {Biomedical
  question-focused multi-document summarization : ILSP and AUEB at BioASQ3}.
  In: CLEF 2015 Working Notes (2015)

\bibitem{Molla:bioasq2017}
Moll{\'{a}}, D.: {Macquarie University at BioASQ 5b --- Query-based
  Summarisation Techniques for Selecting the Ideal Answers}. In: Proc.
  BioNLP2017 (2017)

\bibitem{Molla:bioasq2018}
Moll{\'{a}}, D.: {Macquarie University at BioASQ 6b: Deep learning and deep
  reinforcement learning for query-based multi-document summarisation}. In:
  Proceedings BioASQ Workshop at EMNLP 2018 (2018)

\bibitem{Nallapati2017}
Nallapati, R., Zhai, F., Zhou, B.: {SummaRuNNer: A Recurrent Neural Network
  based Sequence Model for Extractive Summarization of Documents}. In: AAAI
  2017 (nov 2017)

\bibitem{Kumar:bioasq2018}
{Naresh Kumar}, A., Kesavamoorthy, H., Das, M., Kalwad, P., {Raghavi Chandu},
  K., Mitamura, T., Nyberg, E.: {Ontology-Based Retrieval {\&} Neural
  Approaches for BioASQ Ideal Answer Generation}. In: Proceedings BioASQ
  Workshop at EMNLP 2018. pp. 79--89 (2018)

\bibitem{Rush2015}
Rush, A.M., Chopra, S., Weston, J.: {A Neural Attention Model for Abstractive
  Sentence Summarization}. In: In Proceedings of the Conference on Empirical
  Methods in Natural Language Processing (EMNLP). pp. 379--389. No. September
  (2015)

\bibitem{See2017}
See, A., Liu, P.J., Manning, C.D.: {Get To The Point: Summarization with
  Pointer-Generator Networks}. In: ACL 2017 (2017)

\bibitem{Tsatsaronis:2015}
Tsatsaronis, G., Balikas, G., Malakasiotis, P., Partalas, I., Zschunke, M.,
  Alvers, M.R., Weissenborn, D., Krithara, A., Petridis, S., Polychronopoulos,
  D., Almirantis, Y., Pavlopoulos, J., Baskiotis, N., Gallinari, P.,
  Arti{\'{e}}res, T., Ngomo, A.C.N., Heino, N., Gaussier, E., Barrio-Alvers,
  L., Schroeder, M., Androutsopoulos, I., Paliouras, G.: {An Overview of the
  BIOASQ Large-Scale Biomedical Semantic Indexing and Question Answering
  Competition}. BMC Bioinformatics  \textbf{16}(1), ~138 (2015).
  \doi{10.1186/s12859-015-0564-6}

\bibitem{Zhang2018}
Zhang, X., Lapata, M., Wei, F., Zhou, M.: {Neural Latent Extractive Document
  Summarization}. In: EMNLP 2018 (aug 2018)

\end{thebibliography}
%
%
%
%
%

%
%

\end{document}